\documentclass[letterpaper]{article} 
\usepackage[draft]{aaai2026}  
\usepackage{times}  
\usepackage{helvet}  
\usepackage{courier}  
\usepackage[hyphens]{url}  
\usepackage{graphicx} 
\usepackage{natbib}  
\usepackage{caption} 
\usepackage{algorithm}
\usepackage{algorithmicx}
\usepackage{algpseudocode}

\usepackage{amsmath}
\usepackage{amssymb}
\usepackage{amsthm}
\usepackage{makecell}
\usepackage[table]{xcolor}
\newcommand{\yh}[1]{\textcolor{black}{#1}}
\newcommand{\method}{M3OOD}

\usepackage{newfloat}
\usepackage{listings}
\DeclareCaptionStyle{ruled}{labelfont=normalfont,labelsep=colon,strut=off} 
\floatstyle{ruled}
\newfloat{listing}{tb}{lst}{}
\floatname{listing}{Listing}
\title{M3OOD: Automatic Selection of Multimodal OOD Detectors}
\author{Yuehan Qin, Li Li, Defu Cao, Tiankai Yang, Jiate Li, Yue Zhao}
\affiliations{University of Southern California\\
\{yuehanqi, lli15753, defucao, tiankaiy, jiateli, yzhao010\}@usc.edu
}

\usepackage{bibentry}

\begin{document}
\maketitle

\begin{abstract}
Out-of-distribution (OOD) robustness is a critical challenge for modern machine learning systems, particularly as they increasingly operate in multimodal settings involving inputs like video, audio, and sensor data. Currently, many OOD detection methods have been proposed, each with different designs targeting various distribution shifts. A single OOD detector may not prevail across all the scenarios; therefore, how can we automatically select an ideal OOD detection model for different distribution shifts? Due to the inherent unsupervised nature of the OOD detection task, it is difficult to predict model performance and find a universally Best model. Also, systematically comparing models on the new unseen data is costly or even impractical. To address this challenge, we introduce M3OOD, a meta-learning-based framework for OOD detector selection in multimodal settings. Meta learning offers a solution by learning from historical model behaviors, enabling rapid adaptation to new data distribution shifts with minimal supervision. Our approach combines multimodal embeddings with handcrafted meta-features that capture distributional and cross-modal characteristics to represent datasets. By leveraging historical performance across diverse multimodal benchmarks, M3OOD can recommend suitable detectors for a new data distribution shift. Experimental evaluation demonstrates that M3OOD consistently outperforms 10 competitive baselines across 12 test scenarios with minimal computational overhead.
\end{abstract}

\section{Introduction}
Out-of-distribution (OOD) detection aims to identify samples that differ markedly from the distribution of the training data.
\yh{For multimodal machine learning (ML) systems handling diverse modality inputs like vision, text, and audio, this capability is essential for maintaining robustness \citep{yang2021generalized, song2024better, song2025discovering, hu2024rethinking, wang2024template}.}
and is particularly important in high‑risk domains such as autonomous driving \citep{Li_Ji_Wu_Li_Qin_Wei_Zimmermann_2024}, medical diagnostics \citep{ulmer2020trust}, and other applications \citep{xu2025legolearnlabelefficientgraphopenset, xu2025fewshotgraphoutofdistributiondetection, xu2025glipoodzeroshotgraphood,sun2023efficient, song2022adaptive, song2025coact, shi2025efficient, song2023nlpbench,li2023biased}.
As ML systems are increasingly adopted in multimodal settings \citep{radford2021learningtransferablevisualmodels,zhang2023vpgtranstransfervisualprompt,li2024domain,li2025secure,li2025treble}, researchers have begun to explore specialized benchmarks and frameworks for multimodal OOD detection \citep{dong2024multiood,li2024dpudynamicprototypeupdating}.
However, although a broad range of OOD detection methods have been proposed, each tuned to capture particular characteristics of data distributions, there is still no systematic approach for selecting the most suitable OOD detector under multimodal settings.
This challenge stems from the inherently unsupervised nature of the OOD detection, which makes predicting model performance and identifying a universally optimal model challenging.
Given that each OOD detector is based on distinct assumptions and methodological choices, selecting a single model for all distributions is ineffective, and exhaustively training one per case is infeasible.
This is even compounded by the fact that cross-modal alignment inconsistencies and modality-specific distribution shifts can cause methods that perform well on individual modalities to fail when modalities are combined, consistent with the no‑free‑lunch theorem \citep{wolpert1997no}. 
Furthermore, conducting systematic model comparisons on the new data distribution shifts can be prohibitively expensive or unfeasible. 
As a result, we require an automated framework that can identify the most appropriate OOD detector without expensive evaluations.

\textbf{Present Work}. 
In the related field of outlier detection (OD), unsupervised OD method selection has advanced through meta‑learning as in MetaOD \citep{NEURIPS2021_23c89427}, ELECT \citep{10027699}, ADGym \citep{jiang2024adgym}, and HyPer \citep{ding2024hyper}. However, these approaches cannot be directly applied to OOD detection due to differences in task structure and data complexity.
OOD detection is trained only on in-distribution (ID) data to identify samples from entirely different distributions, often across multiple datasets.
It often deals with high‑dimensional visual data, in contrast to OD, which is commonly applied to time‑series or tabular datasets.
Extending from unimodal OOD data \citep{qin2025metaoodautomaticselectionood} to multimodal inputs introduces additional challenges: adapting embeddings to handle temporal sequences and spatial information, maintaining efficiency as dataset size grows, and creating unified representations for heterogeneous data types.
More critically, designing effective multimodal meta-features for model selection becomes more difficult due to the complex interactions between modalities and the lack of clear guidelines for capturing cross-modal relationships in feature representations.
Differences in data representation, distributional characteristics, and detection behavior across modalities require a specialized approach for multimodal OOD detection model selection.

\textbf{Our Work}.
To address the aforementioned challenges,  we present \method{} (Fig.~\ref{fig:pipeline}),
the \textit{first} 
model selection approach for OOD detection in multimodal settings, based on meta-learning.
We show that by combining multimodal model embeddings with handcrafted meta-features capturing distributional and modality-specific properties,
our meta-learning based approach can unify representations for video and optical flow.
Alterations to a single modality can be reflected in the meta-embeddings, enabling the model to adapt its detector selection accordingly.
The central idea is that an OOD detector that performs well on previous datasets with similar properties is likely to generalize well to new datasets. During the meta-training phase, we evaluate a pool of OOD detection methods across a wide range of carefully curated datasets spanning different modalities, including videos and spectrograms, to build a performance profile under varied conditions. When a new multimodal dataset is introduced, we utilize the knowledge accumulated from historical datasets to recommend an appropriate OOD detection method. This selection process is guided by estimating the similarity between the new dataset and those seen during meta-training.
Our main
contributions are:

\begin{itemize}
   \item \textbf{First Multimodal OOD Detection Model Selection Framework}. To our best knowledge, we introduce the first meta-learning-based framework for zero-shot multimodal OOD detection model selection.
   \item \textbf{Specialized Multimodal Embeddings}. We leverage multimodal features to quantify similarity among OOD detection tasks, enhancing detector selection through better characterization of cross-modal OOD properties.
   \item \textbf{Superior Performance}.
   \method{} outperforms eleven model selection methods and unsupervised meta-learners across twelve test data pairs,
   yielding statistically significant ranking improvements with efficient runtime.
   \item \textbf{Open-Source Release}. We provide the testbed and code
   at https://github.com/yqin43/M3OOD.
\end{itemize}

\section{Related Work}

\subsection{Multimodal OOD Detection}

Multimodal OOD detection has gained attention in recent work, particularly for vision-language architectures \citep{ming2022delvingoutofdistributiondetectionvisionlanguage,wang2023clipnzeroshotooddetection}. Maximum Concept Matching (MCM) \citep{ming2022delvingoutofdistributiondetectionvisionlanguage} leverages alignment between visual features and textual concept representations to generate OOD scores. CLIPN \citep{wang2023clipnzeroshotooddetection} enhances the CLIP architecture through contrasting prompt techniques that strengthen the distinction between ID and OOD data. Furthermore, comprehensive multimodal benchmarks incorporating video, optical flow, and audio modalities have been developed
\citep{dong2024multioodscalingoutofdistributiondetection, li2024dpudynamicprototypeupdating}, highlighting cross-modal prediction inconsistencies in OOD scenarios.

\subsection{Unsupervised OOD Detector Selection}
A core difficulty in OOD detection stems from the inherent unknowns: the characteristics of OOD samples and the properties of their underlying distributions remain hidden during the training phase \citep{hendrycks2019benchmark, liang2018enhancing}.
This constraint forces model selection to operate in an unsupervised manner, as labeled OOD samples are unavailable for evaluating and comparing detector performance \citep{lee2018simple}.
In real-world deployment scenarios, OOD detectors must handle test inputs spanning diverse and previously unseen distributional patterns \citep{yang2021generalized}, making unsupervised model selection approaches particularly advantageous for OOD detection applications \citep{liu2020energy}. While recent work has explored instance-specific selection of detector ensembles \citep{xue2024enhancing}, such methods differ from our objective of identifying a single optimal detector for each dataset.

Many existing approaches to OOD detector selection rely on trial-and-error or empirical heuristics. A straightforward strategy is to default to widely used detectors such as Maximum Softmax Probability \citep{hendrycks17baseline} or ODIN \citep{Liang2017EnhancingTR}. Other simple methods include using the confidence scores from ID data as proxies for selecting OOD detectors. However,
these heuristic strategies often yield suboptimal performance due to neural networks' tendency toward overconfident OOD predictions \citep{hendrycks17baseline}.
Another line of work explores similarity-based techniques, where model selection is guided by the resemblance between datasets or their clustering, a strategy previously applied in algorithm recommendation systems \citep{conf/ecai/KadiogluMST10,nikolic2013simple,xu2012satzilla2012,journals/ai/MisirS17}. We thus include these methods as baselines in our study.

\subsection{Embedding-based Representations for Meta-learning}
In meta-learning, effective data representation is essential for capturing dataset or task similarity, and embeddings serve as a key mechanism for this purpose. Traditionally, computational meta-features such as dataset statistics and model-independent properties have been widely used to represent data in meta-learning frameworks \citep{vanschoren2018meta}. More recently, advanced learning-based representations which aim to learn embeddings from data directly have emerged, including methods like dataset2vec \citep{jomaa2021dataset2vec} and HyPer \citep{ding2024hyper}. In parallel, language model and multimodal embeddings have been increasingly employed to encode dataset characteristics, offering a semantic-rich alternative that supports deeper model understanding \citep{drori2019automlusingmetadatalanguage, fang2024largelanguagemodelsllmstabular, qin2025metaoodautomaticselectionood}.
To leverage the strengths of both approaches, we combine handcrafted meta-features that capture distributional and multimodality-specific characteristics with SlowFast-generated embeddings for comprehensive multimodal dataset representation.

\section{Methodology}\label{sec:method}
\subsection{Preliminaries on OOD Detection} 
Consider a training dataset $\mathbf{X}_{\text{train}}$ sampled from an in-distribution $\mathcal{P}_{\text{in}}$, such that $\mathbf{X}_{\text{train}} \sim \mathcal{P}_{\text{in}}$.  
Typically, a model is trained solely on ID data to learn a task such as classification. At test time, we consider a dataset $\mathbf{X}_{\text{test}}$ that may contain both ID samples and samples from unknown distributions (OOD samples).  
The goal of OOD detection is to build a classifier $G$ that decides for each test input $x \in \mathbf{X}_{\text{test}}$ whether it originates from $\mathcal{P}_{\text{in}}$:
\[
G(x) =
\begin{cases}
ID & \text{if } x \in \mathcal{P}_{\text{in}},\\
OOD & \text{if } x \notin \mathcal{P}_{\text{in}}.
\end{cases}
\]

When evaluating an OOD detector on $\mathbf{X}_{\text{test}}$, both ID and
OOD test data are present to assess the detector’s capability
to accurately distinguish between known and unknown data
samples. Accordingly, in this study, 
we use dataset pairs $D = \{\mathbf{X}_{\text{train}}, \mathbf{X}_{\text{test}}\}$, where $\mathbf{X}_{\text{train}}$ contains only ID samples and $\mathbf{X}_{\text{test}}$ contains a mixture of ID and OOD samples, allowing us to assess how well the detector separates known from unknown data. When the input data includes multiple modalities, we extend this framework to multimodal OOD detection, which we define as follows:

\subsubsection{MultiModal OOD Detection}
Each training sample $x_i$ contains $K$ distinct modalities, expressed as 
$x_i = \{x_i^{k} \mid k = 1,\ldots,K \}$. 
Information from all these modalities is integrated to generate the final prediction by takeing the combined embeddings from all modalities and outputs a score $s$. $s$ may represent a probability, a confidence score, an energy value, or any other scalar used by the detector.
Let $\psi(\cdot)$ be a feature extractor that maps an input $x_i$ to an embedding $E$,
and let $h(\cdot)$ be a scoring function or classifier that maps this embedding to an output score $s$.
The overall output used for OOD detection is:
\[
s = h\!\big(\psi(x_i)\big) = h([\psi(x_1), ..., \psi(x_K)]) = h([E_1, ..., E_K]),
\]
A sample with score above the threshold $\eta$ is classified as ID; otherwise, it is classified as OOD:
\[
G(x) =
\begin{cases}
ID & \text{if } s >= \eta,\\
OOD & \text{if } s < \eta.
\end{cases}
\]

\subsection{Problem Statement and Framework Overview}
\label{subsec:overview}
Given a new and previously unseen pair of datasets $D_\text{new}= \{\mathbf{X}_{\text{train}}^\text{new}, \mathbf{X}_{\text{test}}^\text{new}\}$ as input, our objective is to choose the best candidate OOD detection model $M \in \mathcal{M}$ \textit{without} conducting test-time model evaluations, where we have no ground truth labels $\mathbf{y}_{\text{test}}^\text{new}$ for evaluation. In this work, we adopt a meta-learning approach to transfer performance knowledge from previously encountered tasks to the new OOD detection setting. Meta-learning \citep{vanschoren2018meta}, often referred to as "learning to learn", involves training across a range of historical or meta-level tasks, enabling the algorithm to generalize effectively to novel tasks. The underlying intuition is that an OOD detector that performs well on past datasets with similar characteristics is also likely to perform well on a new, related dataset. This strategy is especially beneficial in situations where model evaluation is impractical or costly due to the absence of ground truth labels or the need for quick deployment.

    The proposed meta-learner, \method{}, relies on:
    
(1) A collection of $n$ historical (i.e., meta-train) OOD detection dataset pairs, $\mathcal{D}_\text{train} = \{D_{1}, \ldots ,D_{n}\}$ with \underline{ground truth labels},
    i.e., ${D}=\{\mathbf{X}_{\text{train}}, (\mathbf{X}_{\text{test}}, \underline{\mathbf{y}_{\text{test}}})\}$ where $\mathbf{X} = [\mathbf{X}^1, \ldots, \mathbf{X}^K]$ denotes $K$-modality data.
    
(2) Historical performance $\mathbf{P}$ of the pre-set model set $\mathcal{M}= \{M_1, \ldots, M_m\}$ (with $m$ models), on the meta-train datasets. 
    We refer to \( \mathbf{P} \in \mathbb{R}^{n \times m} \) as the performance matrix, where 
    \( \mathbf{P}_{i,j} \) corresponds to the \( j \)-th model \( M_j \)'s performance on the 
    \( i \)-th meta-train dataset pair \( D_i \).

\begin{figure*}[h]
    \centering
\includegraphics[width=\linewidth]{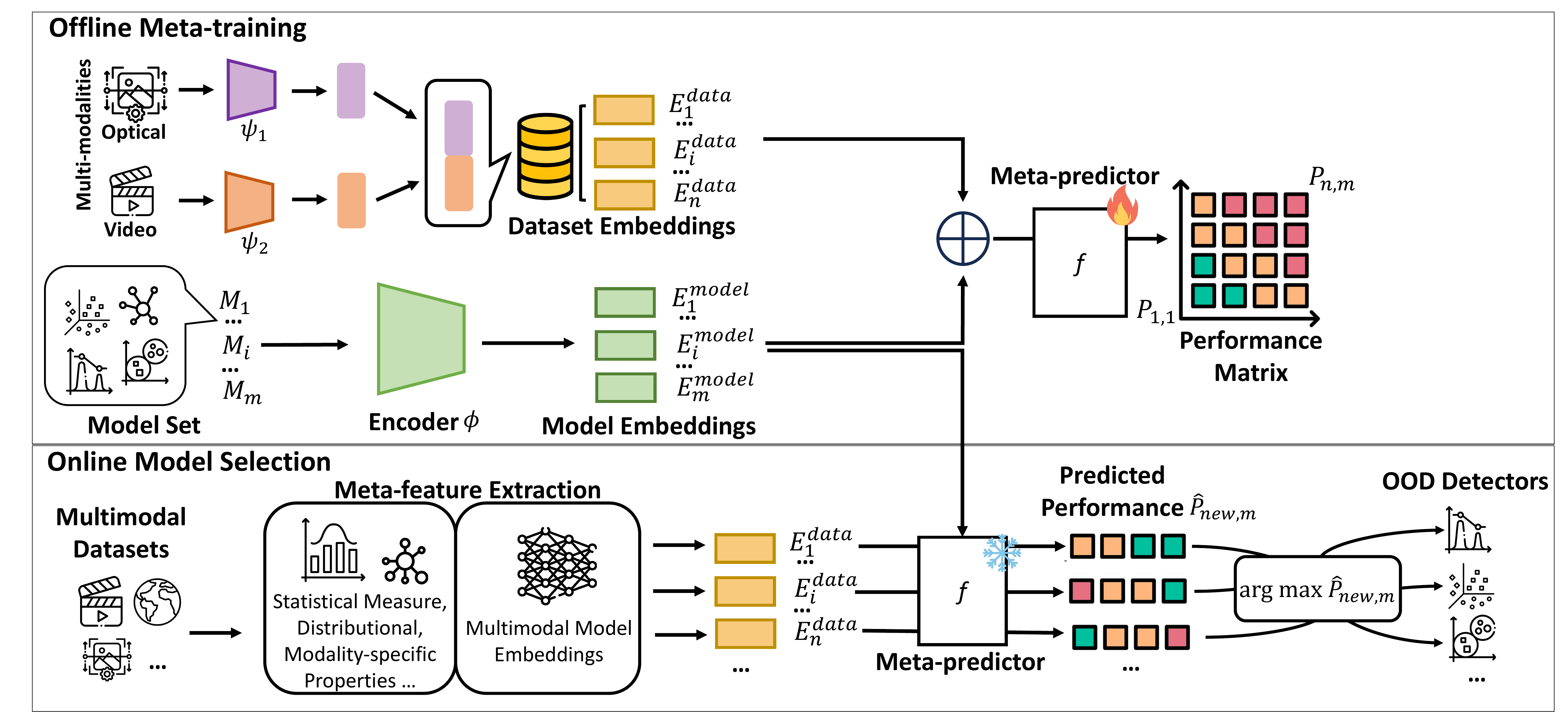}
    \caption{\method{} overview; offline meta-training
  phase is shown on the top: the key is to train 
  a meta performance predictor $f$ to map language embeddings of the datasets and models to their performance $\mathbf{P}$; the online model selection is shown at the bottom by transferring the meta-predictor $f$ to predict the test data paired with OOD detectors for selection. 
  }
    \label{fig:pipeline}
\end{figure*}

Our method includes two stages: (\textit{i}) an offline training stage, where a model is trained to capture how different OOD detection models perform across a set of historical datasets $\mathcal{D}_\text{train}$, and (\textit{ii}) an online stage, where this prior information is used to select an appropriate model for a new test dataset $D_\text{new}$.
Fig.~\ref{fig:pipeline} outlines the workflow and key elements of \method{},
with the offline training phase shown at top and the online model selection stage shown at the bottom.

\subsection{Offline Meta-Training}
\label{subsec:meta-train}
In the offline training stage (Fig.~\ref{fig:pipeline} top), we construct embeddings that represent each combination of a dataset pair $\mathcal{D}_\text{train}$
and a method $\mathcal{M}$, and then learn a latent function that maps these embeddings to their corresponding performance values $\mathbf{P}$. By capturing this relationship, the meta-learner can generalize to new, unseen datasets and identify the method expected to perform best based on the learned mapping ${\mathcal{D}_\text{train}, \mathcal{M}} \rightarrow \mathbf{P}$.
Note this process is supervised.

To predict the performance of the \textit{candidate model} on \textit{a new dataset pair}, we propose training a meta-predictor as a regression problem. 
The input to the meta-predictor consists of ${E_{i}^\text{meta}, E_{j}^\text{model}}$, corresponding to the embedding
of the $i$-th dataset pair and the embedding of the $j$-th OOD detector.
Dataset embedding of dataset pair $\mathcal{D}$ is denoted as $E^\text{data} = \psi(\mathcal{D})$, and method embedding extracted from $\mathcal{M}$ is denoted as $E^\text{model} = \phi(\mathcal{M}, M)$;
we provide more details below on the embedding generation in the following session.
The meta-train process can be formulated as:
\begin{equation} \label{eq:1}
\begin{aligned}
f : \mathcal{H} \times \mathcal{G} \rightarrow \mathbb{R}^+ \\
\text{where } \mathcal{H} &= \{\bigl[E_{i}^\text{video},\, E_{i}^\text{flow}\bigr] \mid i \in \{1,\dots,n\}\} \\
\mathcal{G} &= \{E_{j}^\text{model} \mid j \in \{1,\dots,m\}\} \\
f(E_i^{dataset}, E_j^{model}) &= \hat{P}_{i,j} \quad \forall (i,j) \in [n] \times [m]
\end{aligned}
\end{equation}
Our goal is to train the meta-predictor $f$\footnote{The format of $f$ can be any regression models; in this work, we use an XGBoost \citep{chen2016xgboost} model due to its balance of simplicity and expressiveness, as well as strong feature selection characteristic.} to map the characteristics of the datasets and the OOD detectors to their corresponding performance ranking across all historical dataset pairs. Specifically, the meta-train objective is as follows:
\begin{equation} \label{eq:obj}
\min_{f} \mathbb{E}_{i \sim \mathcal{N}, j \sim \mathcal{M}} \left[ \mathcal{L}(f(\psi(D_i), \phi(\mathcal{M}, M_j)), P_{i,j}) \right]
\end{equation}

\subsection{Data and Model Embeddings}\label{two_embed}
A central component of our meta-learning framework involves extracting meta-features that characterize the essential properties of diverse datasets. Different OOD detection models employ varying algorithmic principles (e.g., probability-based, logit-based, feature-based) and operate under distinct assumptions about distribution shifts and OOD patterns. Consequently, model performance varies significantly based on the underlying dataset characteristics and types of samples present. When encountering a new task, the objective is to identify datasets in the meta-training repository that exhibit comparable properties and leverage models that have demonstrated strong performance on those similar tasks.
To achieve this, the data and model embeddings, used as inputs to $f$, need to provide concise and consistent representations of the datasets and models involved, rather than relying on raw data with varying sizes. This requires creating embeddings that capture key characteristics of the data, enabling the model to adapt efficiently to new tasks.
In this study, we examine the use of multimodal model embeddings in conjunction with traditional meta-features that capture distributional and multimodal characteristics.

\subsubsection{Multimodal Model Embeddings} Multimodal models integrate information from multiple input sources by learning joint representations that capture cross-modal relationships and dependencies. In \method{}, we use the SlowFast network \citep{feichtenhofer2019slowfastnetworksvideorecognition} initialized with pre-trained weights from Kinetics-400 \citep{kay2017kineticshumanactionvideo} as our feature extractor to embed the visual information. 
For the optical flow encoder, we adopt the SlowFast architecture with only the slow pathway, using pre-trained weights from Kinetics-400 dataset \citep{kay2017kineticshumanactionvideo}.
We concatenate the embeddings of all modalities
and treat them as a unified entity.

\subsubsection{Meta-features}
Meta-learning fundamentally relies on transferring knowledge from previous tasks to improve performance on new ones, which is only effective when the new task shares structural similarities with historical tasks. The key challenge lies in defining robust representations of task similarity to identify which prior experiences are most relevant for the current problem.
In meta-learning
and feature engineering contexts, similarity between meta-train and test datasets are quantified through characteristic features of a dataset, also known as meta-features \citep{vanschoren2018meta}.

These features span from basic distributional characteristics like variance, skewness, covariance, etc. to multimodality descriptors that capture video temporal patterns and optical flow characteristics (e.g. colourfulness index, edge density, motion characteristics). Meta-features support task understanding by encoding both distributional properties, which are frequently employed in automated machine learning research, and modality-specific patterns, enabling model selection for new multimodal data through analogies with previously encountered tasks.
The complete list of these features are in Supplementary Material \S 2.

\subsection{Online Model Selection}
\label{subsec:model-selection}
During online model selection (Fig.~\ref{fig:pipeline} bottom), we generate embeddings for the test dataset pair $D_\text{new}$, reuse the precomputed embeddings of each model $\mathcal{M}$, and use the meta performance predictor $f$ trained during the offline phase, to estimate the performance of different OOD detection models. The model with the highest predicted score is then selected, following the procedure in Eq. (\ref{eq:goal}).

\begin{equation}
    M^* := \underset{M_j \in \mathcal{M}}{\arg\max} \, \widehat{\mathbf{P}}_{\text{new}, j}, \quad \text{where} \quad \widehat{\mathbf{P}}_{\text{new}, j} = f(E^{\text{meta}}_{\text{new}}, E^{\text{model}}_j)
    \label{eq:goal}
\end{equation}

For a new dataset pair, the trained predictor $f$ is used to estimate the relative performance of various OOD detectors. Based on these predictions, the method ranked highest is selected\footnote{Although selecting the top-$k$ methods for ensemble use is possible, this work focuses on top-1 selection.}, as indicated in Eq. (\ref{eq:goal}). Notably, this approach is \textit{zero-shot}, meaning it does not involve any model training on the test data.

\section{Experiments}
Our experiments address the following research questions (RQ):
\textbf{RQ1}:
How effective is the proposed \method{} in unsupervised OOD detector selection compared to leading baselines? 
\textbf{RQ2}:
How do different design choices affect the perofmance of \method{}?\textbf{RQ3}
: How much time overhead/saving \method{} introduces to multimodal OOD detection?

\textbf{The model set $\mathcal{M}$}. 
We construct $\mathcal{M}$ with 9 popular OOD detectors (Tab.~\ref{table:ood_model}) spanning diverse detection methods.

\begin{table}
\centering
\small
\begin{tabular}{cc} 
\hline
\textbf{Category} & \textbf{OOD Detection Model} \\
\hline
Probability-based & MSP \citep{hendrycks17baseline}\\
& GEN \citep{10203747} \\
\hline
Logit-based & MaxLogit \citep{Hendrycks2022ScalingOD} \\
& EnergyBased \citep{NEURIPS2020_f5496252} \\
\hline
Feature-based & Mahalanobis \citep{NEURIPS2018_abdeb6f5} \\
& ViM \citep{haoqi2022vim} \\
& $k$NN \citep{1053964}\\
\hline
Activation Pruning
& ReAct \citep{sun2021reactoutofdistributiondetectionrectified}\\
& ASH \citep{djurisic2023extremelysimpleactivationshaping}\\
\hline
\end{tabular}
\caption{OOD detectors considered for model selection.}
\label{table:ood_model}
\end{table}

\textbf{Datasets}.
In real-world settings, OOD data often differ from ID data not only in semantics but also in domain. To better reflect such situation, we design the dataset to include two types of distribution shifts: Far-OOD and Near-OOD. In the Far-OOD setting, we treat a full dataset as ID and use other datasets with related tasks but no overlapping categories as OOD. This introduces both semantic and domain shifts between ID and OOD samples. In the Near-OOD setting, we split the categories within a single dataset into two disjoint groups: one used as ID and the other as OOD. In this case, ID and OOD samples share the same underlying distribution, differing only in semantics.
We use five action recognition datasets (EPIC-Kitchens \citep{munro2020multimodaldomainadaptationfinegrained}, HAC \citep{dong2023simmmdgsimpleeffectiveframework}, HMDB51 \citep{6126543}, UCF101 \citep{soomro2012ucf101dataset101human}, and Kinetics-600 \citep{carreira2018shortnotekinetics600}),
The Near-OOD and Far-OOD setup details are in Supplementary Material \S 4.

\textbf{Train-test Split}.
We split the train-test sets as shown in Tab.~\ref{tab:dataset_splits} in the meta-training stage. Each row corresponds to a different meta-train/meta-test split. For example, in the first row, the meta-train set includes HMDB and Kinetics (Near-OOD and Far-OOD), and the corresponding meta-test set includes UCF and EPIC (Near-OOD). This setup ensures that the meta-predictor is trained on diverse OOD conditions and evaluated on unseen datasets, allowing us to assess its ability to generalize OOD detector selection across both semantic similarity and distributional shifts.
\begin{table*}[h]
\centering
\begin{tabular}{@{}ll@{}}
\hline
\textbf{Train} & \textbf{Test} \\
\hline
HMDB, Kinetics, HMDB-Far-OOD, Kinetics-Far-OOD & UCF, EPIC \\
UCF, EPIC, Kinetics, Kinetics-Far-OOD & HMDB, HMDB-Far-OOD \\
HMDB, UCF, EPIC, HMDB-Far-OOD & Kinetics, Kinetics-Far-OOD \\
\hline
\end{tabular}
\caption{Meta-train train/test split (datasets without the “-Far-OOD” suffix are Near-OOD).}
\label{tab:dataset_splits}
\end{table*}

\textbf{Hardware}.
All models are implemented on the MultiOOD codebase \citep{dong2024multioodscalingoutofdistributiondetection} and run on a multi-NVIDIA RTX 6000 Ada workstation.

\textbf{Training the meta-predictor $f$} (see details in previous Offline Meta-Training section).
In this work, we use an XGBoost \citep{chen2016xgboost} model as $f$ due to its simplicity and expressiveness.

\textbf{Evaluation}.
To evaluate the performance of \method{} against the baselines, we compare the rank of performance of the OOD detector selected by each method among all candidates. We use Area Under the ROC Curve (AUC-ROC) as the evaluation metric\footnote{Other metrics can be used at interest.}, and visualize the results using boxplots and a rank diagram that reports the average rank across all dataset pairs. A rank of 1 indicates the best-performing selection, 11 is the worst (10 baselines plus \method{}). For statistical comparison, we apply the pairwise Wilcoxon rank-sum test across dataset pairs with a significance threshold of $p < 0.05$.

\begin{table}[h]
\centering
\small
\caption{
Various OOD detectors’ performances on Near-OOD and Far-OOD dataset pairs. We highlight the selected OOD method for each dataset pair in the test set in bold.}
\begin{tabular}{lcccc}
\hline
\textbf{OOD Dataset} & HMDB51 & UCF101 & EPIC & Kinetics-600 \\
\hline
\textbf{Near-OOD} &&&& \\
MSP         & 87.74 & 95.73 & 67.59 & 76.16 \\
Energy      & 87.46 & 96.06 & 68.29 & 75.49 \\
MaxLogit    & 87.75 & 96.02 & 68.29 & 75.98 \\
Mahalanobis & 85.28 & 97.14 & 42.99 & 35.83 \\
ReAct       & 87.09 & 95.85 & 65.89 & 73.80 \\
ASH         & 87.16 & 94.02 & 67.92 & 76.16 \\
GEN         & 87.49 & 95.64 & \textbf{68.52} & 75.33 \\
KNN         & \textbf{88.46} & 96.93 & 63.60 & 74.64 \\
VIM         & 88.06 & \textbf{97.66} & 65.60 & \textbf{75.47} \\
\hline
\textbf{Far-OOD} &&&& \\
ID (HMDB51) & Kinetics & UCF& EPIC& HAC\\
\hline
MSP         & 92.48 & 87.95 & 89.07 & 92.28 \\
Energy      & 87.81 & 84.22 & 92.22 & 90.23 \\
MaxLogit    & 90.34 & 87.91 & 91.88 & 91.99 \\
Mahalanobis & 95.01 & 89.34 & 93.66 & 94.56 \\
ReAct       & 97.01 & 91.45 & 98.15 & 95.93 \\
ASH         & 95.35 & 92.41 & 98.46 & 95.39 \\
GEN         & 95.45 & \textbf{93.53} & 99.30 & 95.66 \\
KNN         & 96.70 & 92.33 & 98.97 & 97.26 \\
VIM         & \textbf{98.74} & 94.42 & \textbf{99.63} & \textbf{99.16} \\
\hline
\textbf{Far-OOD} &&&& \\
 ID (Kinetics-600) & HMDB & UCF& EPIC& HAC\\
 \hline
MSP         & 71.75 & 71.49 & 82.05 & 75.07 \\
Energy      & 76.66 & 72.38 & 88.05 & 80.15 \\
MaxLogit    & 78.43 & 73.97 & 84.90 & 80.30 \\
Mahalanobis & 78.84 & 74.33 & 82.69 & 79.51 \\
ReAct       & 71.88 & 70.55 & 84.98 & 75.15 \\
ASH         & 80.84 & 78.20 & 82.99 & 85.93 \\
GEN         & 83.77 & 84.19 & 83.30 & 88.20 \\
KNN         & 84.30 & 82.54 & 83.47 & 96.58 \\
VIM         & \textbf{81.51} & \textbf{78.38} & \textbf{83.50} & \textbf{99.30} \\
\hline
\end{tabular}
\end{table}

\textbf{Model Selection Baselines}.
\begin{table}[th]
\centering
\begin{tabular}{l}
\hline
\textbf{Model Selection Baselines} \\
\hline
\textbf{No model selection or random selection}\\
MSP~\citep{hendrycks17baseline} \\
Mahalanobis (MD)~\citep{lee2018simple} \\
Mega Ensemble (ME) \\
Random Selection (Random) \\
\hline
\textbf{Simple meta-learners (non-optimization)} \\
Global Best (GB) \\
ISAC~\citep{conf/ecai/KadiogluMST10} \\
ARGOSMART (AS)~\citep{nikolic2013simple} \\
\hline
\textbf{Optimization-based meta-learners} \\
ALORS~\citep{journals/ai/MisirS17} \\
NCF~\citep{10.1145/3038912.3052569} \\
\hline
\textbf{Large language models (LLMs) as model selectors} \\
GPT-4o-mini~\citep{openai2024gpt4technicalreport}\\
\hline
\end{tabular}
\caption{Categories of OOD detector selection method baselines in this study.}
\label{tab:ood_selection_methods}
\end{table}
We select the baselines based on the literature in meta-learning for unsupervised model selection \citep{NEURIPS2021_23c89427,10027699,jiang2024adgym,park2023metagl}, grouped into four categories, as shown in Tab.~\ref{tab:ood_selection_methods}.
Details of the model selection baselines are in Supplementary Material \S 3.

\section{Results}\label{sec:results}
In Fig.~\ref{fig:boxplot}, we report the distribution of the true ranks of the top-1 OOD detector selected by each model selection method across the test data pairs. Also, we include the overall average-rank diagram in Fig.~\ref{fig:cd_diagram}, which displays the mean performance rank of the OOD detector selected by each algorithm. To compare two model selection algorithms (e.g., ours with a baseline), we perform Wilcoxon rank test on the rank of the top-1 models selected by our method and the baseline method, as shown in Fig.~\ref{table:p_val}. We summarize the main findings as below:
\begin{figure}[h]
    \centering
\includegraphics[width=\linewidth]{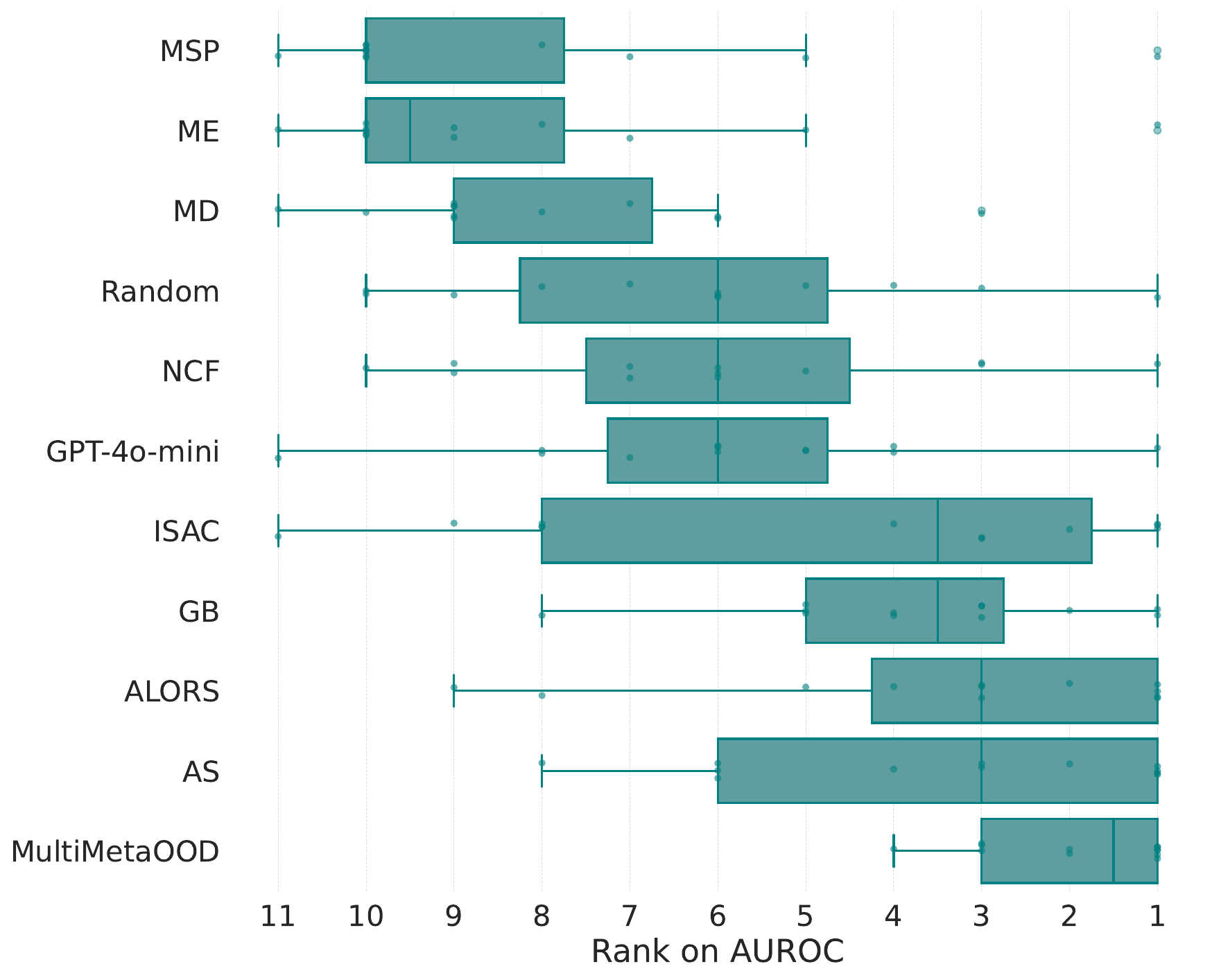}
    \caption{Boxplot of the rank distribution of \method{} and baselines (the lower, the better). \method{} is the lowest/best.}
    \label{fig:boxplot}
\end{figure}
\begin{figure}[h]
    \centering
\includegraphics[width=\linewidth]{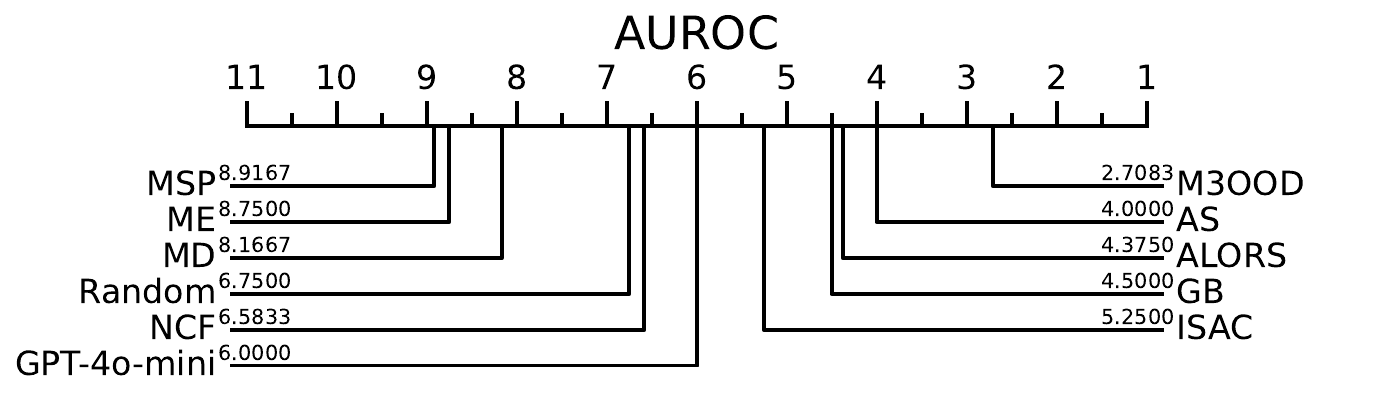}
    \caption{Average rank (lower is better) of methods
w.r.t. performance across datasets; \method{} outperforms all baselines with the lowest rank.}
    \label{fig:cd_diagram}
\end{figure}

\begin{table}
\centering
\begin{tabular}{ccc} 
\hline
 \textbf{Ours} & \textbf{Baseline} & \textbf{p-value}\\
 \hline
 & ME & 0.001 \\
 & \textbf{AS} & \textbf{0.0625} \\
 & ISAC & 0.0156\\
 & \textbf{ALORS} & \textbf{0.125}\\
\textbf{\method{}} & Random & 0.0029 \\
 & MSP & 0.001 \\
 & MD & 0.0005 \\
 & NCF & 0.0039\\
 & GB & 0.0312\\
 & GPT-4o mini & 0.0015\\
 \hline
\end{tabular}
\captionof{table}{Wilcoxon signed-rank test result (marked bold if no significance). \method{} is statistically better than all the baselines except AS and ALORS.}
\label{table:p_val}
\end{table}

\textbf{1. \method{} outperforms all baselines}.
Fig.~\ref{fig:boxplot} demonstrates that \method{} achieves stable, high-quality performance with minimal variance. It maintains the highest average ranking among all the 10 baseline methods that span from random or fixed selection to optimization or learning-based methods (Fig.~\ref{fig:cd_diagram}). Additionally, Tab.~\ref{table:p_val} shows that most performance gains are statistically significant. This consistent pattern of results indicates that \method{} effectively handles complex datasets while maintaining stable. We attribute this effectiveness to the integration of a meta-learning approach with our designed multimodal dataset embeddings.

\textbf{2. Meta-learner perform better than other baselines}. Meta-learners (\method{}, AS, ALORS) significantly outperform single outlier detection methods and ME that averages all the model performances. Meanwhile, optimization-based meta learners (i.e. \method{}, ALORS) demonstrate relatively strong and stable performance in model selection. This improvement comes from two factors. First, meta-learning uses knowledge gained from previous tasks to adapt more effectively to new ones, extracting shared patterns and representations that boost generalization. Second, the optimization routines in these methods drive models to efficient, high-quality solutions. By incorporating multimodal embeddings and meta-features, they can map model performance more accurately than simple meta-learners.

\textbf{3. The poor performance of the no-selection and random-selection baselines highlights the need for OOD model selection}.
Simply averaging the OOD detection scores of all models yields subpar results, as shown in Fig.~\ref{fig:cd_diagram} and Fig.~\ref{fig:boxplot}. Some models underperform consistently across datasets, so treating every model equally drags down overall effectiveness. While selective ensembles can help \citep{zhao2019dcso}, building and running many models is often too costly. In contrast, \method{} leverages offline meta-training to learn which single model to choose, avoiding ensemble construction and enabling efficient testing. Moreover, random selection falls short of all meta-learning baselines. This confirms that each meta-learner offers clear gains over random choice, and that picking an OOD detector at random is not advised. In addition, no single OOD detector achieves strong results on every dataset. This is because different OOD detectors target different dataset characteristics, and real-world data vary widely in their properties. Relying on just one method limits the range of solutions and makes it difficult to handle distribution shifts between datasets.

\textbf{4. LLM as zero-shot model-selector does not perform well under multimodal setting.} GPT-4o-mini may not be well-suited for capturing the nuanced relationships between multimodal dataset characteristics and OOD detector selection, which likely requires more specialized understanding of how different modalities interact and how various detectors respond to specific types of distribution shifts. This indicates that while LLMs offer accessibility, specialized meta-learning methods still hold
substantial
advantages for complex, heterogeneous-input settings such as choosing detectors for multimodal OOD detection.

\subsection{Ablation Studies and Additional Analysis}
\subsubsection{Choice of Meta Predictor}
\begin{figure}[h]
    \centering
\includegraphics[width=\linewidth]{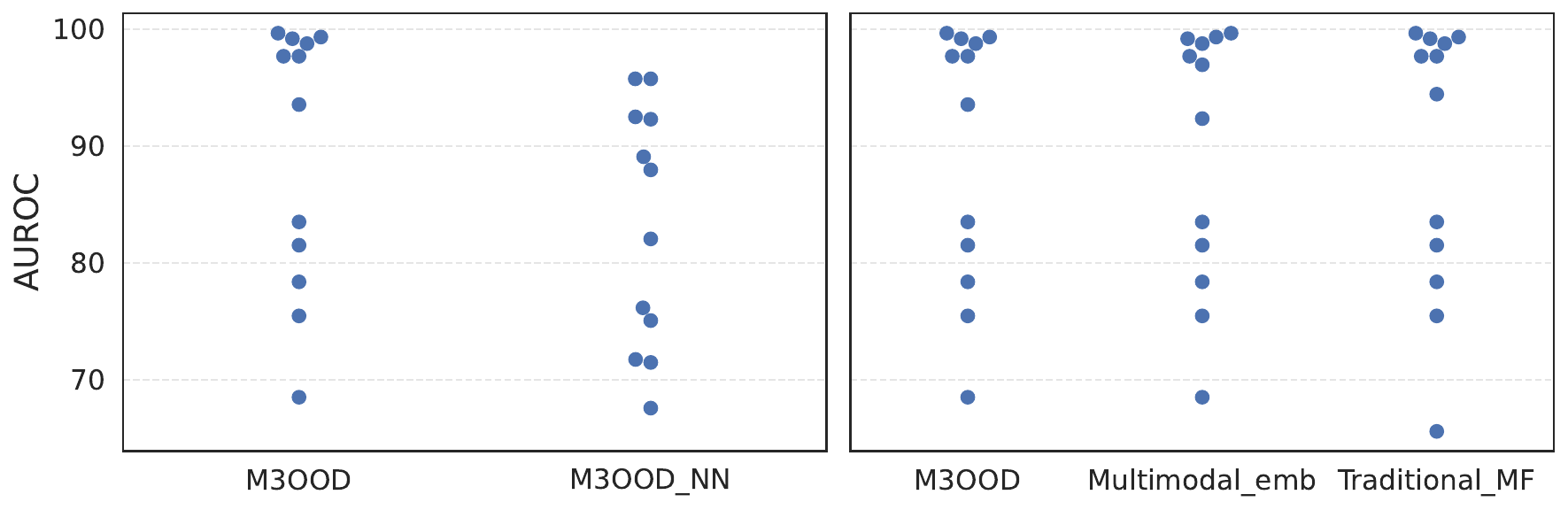}
    \caption{Left: ablation study on different choices of meta-predictor $f$. Tree-based models have better performance. Right: ablation study on different meta embeddings. \method{} has better performance over its variants.}
    \label{fig:ab}
\end{figure}

We evaluate the performances of different choice of meta-predictor $f$. We test \method{} with $f$ replaced by a two-layer MLP meta predictor. The result is shown in Fig.~\ref{fig:ab} left.
Similar to what Jiang et al. (2024) observed in their study, XGBoost and other tree-based models produce more reliable and better-performing meta-predictors than neural network alternatives.

\subsubsection{Choice of Meta Embedding}
We compare the performance of \method{} altered meta-embedding inputs: one that excludes traditional meta-features (Multimoda\_emb), and another that excludes multimodal model embeddings (Traditional\_mf). Fig.~\ref{fig:ab} right shows that combining multimodal model embeddings with traditional meta-features leads to improved performance. This suggests that the two types of meta-information provide complementary signals for selecting effective OOD detectors.

\subsubsection{Runtime Analysis}
\begin{figure}[h]
    \centering
\includegraphics[width=\linewidth]{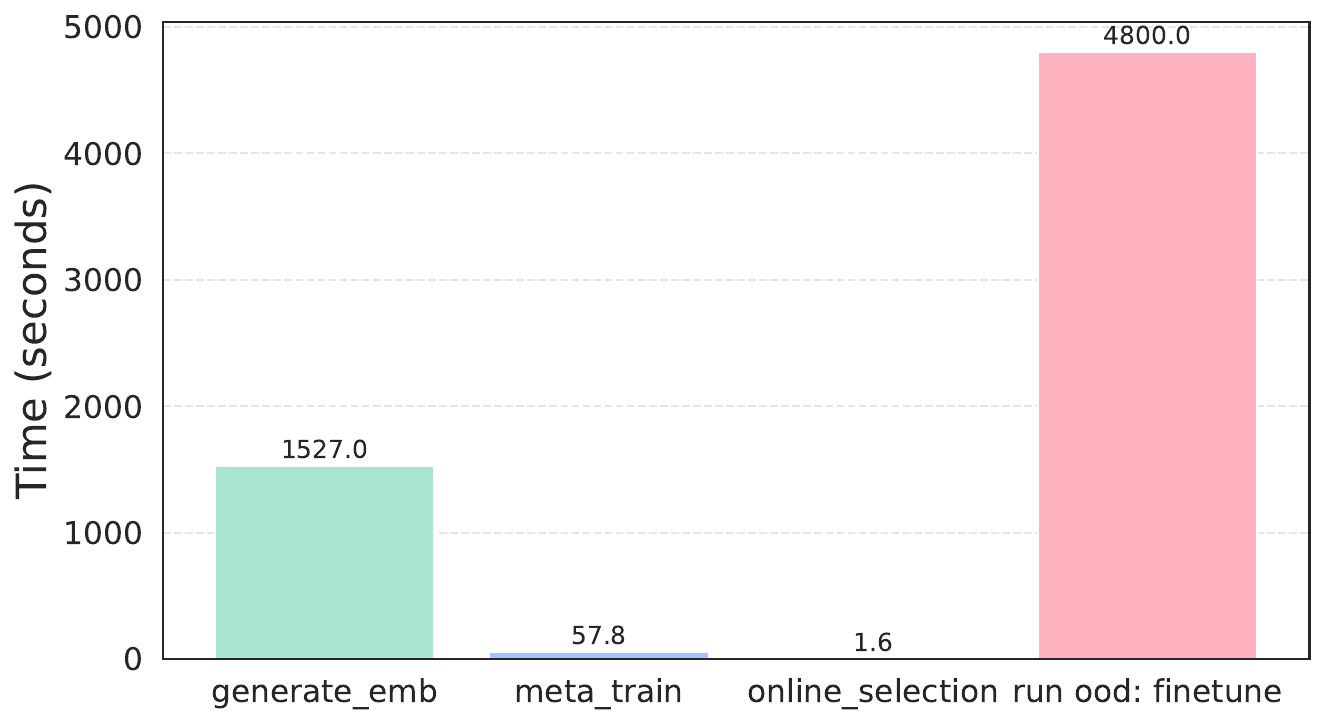}
    \caption{Runtime of \method{} components vs. time required for multimodal OOD detection on the HMDB dataset.
  \method{} incurs a small overhead.}
    \label{fig:runtime}
\end{figure}

OOD detection on large datasets is computationally expensive. Fig.~\ref{fig:runtime} compares the runtime of \method{} components against direct OOD detector execution on the HMDB dataset. While direct OOD detection requires extensive finetuning (HMDB requires 2 mins/ epoch for $\sim$40 epochs, while Kinetics needs 10 mins/ epoch for $\sim$40 epochs), \method{} incurs minimal overhead with embedding generation (1577 seconds), meta-learning (57.8 seconds), and online selection (1.6 seconds) for the HMDB dataset. This demonstrates that our model selection approach achieves significant computational savings compared to running OOD detectors directly.

\section{Conclusion}
In this paper, we propose \method{}, the \textit{first} framework for multimodal OOD detector selection. Our meta-learner draws on a large set of historical OOD detector and dataset-pair records, using multimodal-related meta-features to guide model choice by learning from prior results.
However, \method{} requires sufficient, high-quality historical dataset pairs, which can limit its performance when such data are scarce or not closely related.
For future work, we will expand our evaluation suite to cover a broader range of datasets and models, thereby improving \method{}’s meta-learning capabilities.
We also plan to integrate an uncertainty estimation component so that \method{} can return an “I do not know” result when transferable meta-knowledge is insufficient, making it more reliable in challenging scenarios.

\bibliography{aaai2026}

@article{sun2023efficient,
  title={Efficient and robust KPI outlier detection for large-scale datacenters},
  author={Sun, Yongqian and Cheng, Daguo and Yang, Tiankai and Ji, Yuhe and Zhang, Shenglin and Zhu, Man and Xiong, Xiao and Fan, Qiliang and Liang, Minghan and Pei, Dan and others},
  journal={IEEE Transactions on Computers},
  volume={72},
  number={10},
  pages={2858--2871},
  year={2023},
  publisher={IEEE}
}

@article{song2022adaptive,
  title={Adaptive Ranking-based Sample Selection for Weakly supervised Class-imbalanced Text Classification},
  author={Song, Linxin and Zhang, Jieyu and Yang, Tianxiang and Goto, Masayuki},
  journal={EMNLP 2022 (Findings)},
  year={2022}
}

@article{song2023nlpbench,
  title={Nlpbench: Evaluating large language models on solving nlp problems},
  author={Song, Linxin and Zhang, Jieyu and Cheng, Lechao and Zhou, Pengyuan and Zhou, Tianyi and Li, Irene},
  journal={arXiv preprint arXiv:2309.15630},
  year={2023}
}

@article{song2024better,
  title={Better Explain Transformers by Illuminating Important Information},
  author={Song, Linxin and Cui, Yan and Luo, Ao and Lecue, Freddy and Li, Irene},
  journal={EACL 2024 (Findings)},
  year={2024}
}

@article{hu2024rethinking,
  title={Rethinking llm-based preference evaluation},
  author={Hu, Zhengyu and Song, Linxin and Zhang, Jieyu and Xiao, Zheyuan and Wang, Jingang and Chen, Zhenyu and Xiong, Hui},
  journal={arXiv e-prints},
  pages={arXiv--2407},
  year={2024}
}

@article{wang2024template,
  title={Template Matters: Understanding the Role of Instruction Templates in Multimodal Language Model Evaluation and Training},
  author={Wang*, Shijian and Song*, Linxin and Zhang, Jieyu and Shimizu, Ryotaro and Luo, Ao and Yao, Li and Chen, Cunjian and McAuley, Julian and Wu, Hanqian},
  journal={arXiv preprint arXiv:2412.08307},
  year={2024}
}

@article{song2025discovering,
  title={Discovering knowledge deficiencies of language models on massive knowledge base},
  author={Song, Linxin and Ding, Xuwei and Zhang, Jieyu and Shi, Taiwei and Shimizu, Ryotaro and Gupta, Rahul and Liu, Yang and Kang, Jian and Zhao, Jieyu},
  journal={COLM 2025},
  year={2025}
}

@article{shi2025efficient,
  title={Efficient reinforcement finetuning via adaptive curriculum learning},
  author={Shi, Taiwei and Wu, Yiyang and Song, Linxin and Zhou, Tianyi and Zhao, Jieyu},
  journal={arXiv preprint arXiv:2504.05520},
  year={2025}
}

@article{li2025treble,
  title={Treble counterfactual vlms: A causal approach to hallucination},
  author={Li, Shawn and Qu, Jiashu and Zhou, Yuxiao and Qin, Yuehan and Yang, Tiankai and Zhao, Yue},
  journal={arXiv preprint arXiv:2503.06169},
  year={2025}
}

@article{song2025coact,
  title={CoAct-1: Computer-using Agents with Coding as Actions},
  author={Song, Linxin and Dai, Yutong and Prabhu, Viraj and Zhang, Jieyu and Shi, Taiwei and Li, Li and Li, Junnan and Savarese, Silvio and Chen, Zeyuan and Zhao, Jieyu and others},
  journal={arXiv preprint arXiv:2508.03923},
  year={2025}
}

@String(CVPR= {IEEE Conf. Comput. Vis. Pattern Recog.})

@String(ICASSP=	{ICASSP})

@String(AAAI = {AAAI})

@String(CVPR  = {CVPR})

@String{Computer = "{IEEE} Computer" }

@String{Springer = "Springer-Verlag" }

@inproceedings{NEURIPS2021_23c89427,
 author = {Zhao, Yue and Rossi, Ryan and Akoglu, Leman},
 booktitle = {Advances in Neural Information Processing Systems},
 editor = {M. Ranzato and A. Beygelzimer and Y. Dauphin and P.S. Liang and J. Wortman Vaughan},
 pages = {4489--4502},
 publisher = {Curran Associates, Inc.},
 title = {Automatic Unsupervised Outlier Model Selection},
 url = {https://proceedings.neurips.cc/paper_files/paper/2021/file/23c894276a2c5a16470e6a31f4618d73-Paper.pdf},
 volume = {34},
 year = {2021}
}

@inproceedings{park2023metagl,
  title={Meta{GL}: Evaluation-Free Selection of Graph Learning Models via Meta-Learning},
  author={Namyong Park and Ryan A. Rossi and Nesreen Ahmed and Christos Faloutsos},
  booktitle={The Eleventh International Conference on Learning Representations},
  year={2023},
  url={https://openreview.net/forum?id=C1ns08q9jZ}
}

@inproceedings{10.1145/3038912.3052569,
author = {He, Xiangnan and Liao, Lizi and Zhang, Hanwang and Nie, Liqiang and Hu, Xia and Chua, Tat-Seng},
title = {Neural Collaborative Filtering},
year = {2017},
isbn = {9781450349130},
publisher = {International World Wide Web Conferences Steering Committee},
address = {Republic and Canton of Geneva, CHE},
booktitle = {Proceedings of the 26th International Conference on World Wide Web},
pages = {173–182},
numpages = {10},
location = {Perth, Australia},
series = {WWW '17}
}

@INPROCEEDINGS {10027699,
author = {Y. Zhao and S. Zhang and L. Akoglu},
booktitle = {2022 IEEE International Conference on Data Mining (ICDM)},
title = {Toward Unsupervised Outlier Model Selection},
year = {2022},
volume = {},
issn = {},
pages = {773-782},
publisher = {IEEE Computer Society},
address = {Los Alamitos, CA, USA},
month = {dec}
}

@inproceedings{conf/ecai/KadiogluMST10,
  author = {Kadioglu, Serdar and Malitsky, Yuri and Sellmann, Meinolf and Tierney, Kevin},
  booktitle = {ECAI},
   pages = {751-756},
  publisher = {IOS Press},
  series = {Frontiers in Artificial Intelligence and Applications},
  title = {ISAC - Instance-Specific Algorithm Configuration.},
  url = {http://dblp.uni-trier.de/db/conf/ecai/ecai2010.html#KadiogluMST10},
  volume = 215,
  year = 2010
}

@article{xu2012satzilla2012,
  title={SATzilla2012: Improved algorithm selection based on cost-sensitive classification models},
  author={Xu, Lin and Hutter, Frank and Shen, Jonathan and Hoos, Holger H and Leyton-Brown, Kevin},
  journal={Proceedings of SAT Challenge},
  pages={57--58},
  year={2012}
}

@article{journals/ai/MisirS17,
  author = {Misir, Mustafa and Sebag, Michèle},
  journal = {Artif. Intell.},
  pages = {291-314},
  title = {Alors: An algorithm recommender system.},
  volume = 244,
  year = 2017
}

@article{nikolic2013simple,
  title={Simple algorithm portfolio for SAT},
  author={Nikoli{\'c}, Mladen and Mari{\'c}, Filip and Jani{\v{c}}i{\'c}, Predrag},
  journal={Artificial Intelligence Review},
  volume={40},
  number={4},
  pages={457--465},
  year={2013},
  publisher={Springer}
}

@article{hendrycks17baseline,
  author    = {Dan Hendrycks and Kevin Gimpel},
  title     = {A Baseline for Detecting Misclassified and Out-of-Distribution Examples in Neural Networks},
  journal = {Proceedings of International Conference on Learning Representations},
  year = {2017},
}

@inproceedings{Hendrycks2022ScalingOD,
  title={Scaling Out-of-Distribution Detection for Real-World Settings},
  author={Dan Hendrycks and Steven Basart and Mantas Mazeika and Mohammadreza Mostajabi and Jacob Steinhardt and Dawn Xiaodong Song},
  booktitle={ICML},
  year={2022},
}

@inproceedings{NEURIPS2020_f5496252,
 author = {Liu, Weitang and Wang, Xiaoyun and Owens, John and Li, Yixuan},
 booktitle = {Advances in Neural Information Processing Systems},
 editor = {H. Larochelle and M. Ranzato and R. Hadsell and M.F. Balcan and H. Lin},
 pages = {21464--21475},
 publisher = {Curran Associates, Inc.},
 title = {Energy-based Out-of-distribution Detection},
 volume = {33},
 year = {2020}
}

@article{Liang2017EnhancingTR,
  title={Enhancing The Reliability of Out-of-distribution Image Detection in Neural Networks},
  author={Shiyu Liang and Yixuan Li and Rayadurgam Srikant},
  journal={arXiv: Learning},
  year={2017},
}

@inproceedings{NEURIPS2018_abdeb6f5,
 author = {Lee, Kimin and Lee, Kibok and Lee, Honglak and Shin, Jinwoo},
 booktitle = {Advances in Neural Information Processing Systems},
 editor = {S. Bengio and H. Wallach and H. Larochelle and K. Grauman and N. Cesa-Bianchi and R. Garnett},
 pages = {},
 publisher = {Curran Associates, Inc.},
 title = {A Simple Unified Framework for Detecting Out-of-Distribution Samples and Adversarial Attacks},
 volume = {31},
 year = {2018}
}

@inproceedings{haoqi2022vim,
title = {ViM: Out-Of-Distribution with Virtual-logit Matching},
author = {Wang, Haoqi and Li, Zhizhong and Feng, Litong and Zhang, Wayne},
booktitle = {Proceedings of the IEEE/CVF Conference on Computer Vision and Pattern Recognition},
year = {2022}
}

@ARTICLE{1053964,
  author={Cover, T. and Hart, P.},
  journal={IEEE Transactions on Information Theory}, 
  title={Nearest neighbor pattern classification}, 
  year={1967},
  volume={13},
  number={1},
  pages={21-27},
  }

@inproceedings{xue2024enhancing,
  title={Enhancing the Power of OOD Detection via Sample-Aware Model Selection},
  author={Xue, Feng and He, Zi and Zhang, Yuan and Xie, Chuanlong and Li, Zhenguo and Tan, Falong},
  booktitle={Proceedings of the IEEE/CVF Conference on Computer Vision and Pattern Recognition},
  pages={17148--17157},
  year={2024}
}

@article{jiang2024adgym,
  title={ADGym: Design Choices for Deep Anomaly Detection},
  author={Jiang, Minqi and Hou, Chaochuan and Zheng, Ao and Han, Songqiao and Huang, Hailiang and Wen, Qingsong and Hu, Xiyang and Zhao, Yue},
  journal={Advances in Neural Information Processing Systems},
  volume={36},
  year={2024}
}

@article{yang2021generalized,
  title={Generalized out-of-distribution detection: A survey},
  author={Yang, Jingkang and Zhou, Kaiyang and Li, Yixuan and Liu, Ziwei},
  journal={arXiv preprint arXiv:2110.11334},
  year={2021}
}

@article{wolpert1997no,
  title={No free lunch theorems for optimization},
  author={Wolpert, David H and Macready, William G},
  journal={IEEE transactions on evolutionary computation},
  volume={1},
  number={1},
  pages={67--82},
  year={1997},
  publisher={IEEE}
}

@article{vanschoren2018meta,
  title={Meta-learning: A survey},
  author={Vanschoren, Joaquin},
  journal={arXiv preprint arXiv:1810.03548},
  year={2018}
}

@inproceedings{ding2024hyper,
  title={Fast Unsupervised Deep Outlier Model Selection with Hypernetworks},
  author={Ding, Xueying and Zhao, Yue and Akoglu, Leman},
  booktitle={ACM SIGKDD Conference on Knowledge Discovery and Data Mining},
  publisher={{ACM}},
  year={2024},
}

@inproceedings{ulmer2020trust,
  title={Trust issues: Uncertainty estimation does not enable reliable ood detection on medical tabular data},
  author={Ulmer, Dennis and Meijerink, Lotta and Cin{\`a}, Giovanni},
  booktitle={Machine Learning for Health},
  pages={341--354},
  year={2020},
  organization={PMLR}
}

@inproceedings{hendrycks2019benchmark,
  title={Benchmark for Out-of-Distribution Detection and Robustness},
  author={Hendrycks, Dan and Mazeika, Mantas and Dietterich, Thomas G},
  booktitle={International Conference on Learning Representations},
  year={2019}
}

@inproceedings{liang2018enhancing,
  title={Enhancing the reliability of out-of-distribution image detection in neural networks},
  author={Liang, Shiyu and Li, Yixuan and Srikant, R},
  booktitle={International Conference on Learning Representations},
  year={2018}
}

@inproceedings{lee2018simple,
  title={A Simple Unified Framework for Detecting Out-of-Distribution Samples and Adversarial Attacks},
  author={Lee, Kimin and Lee, Kibok and Lee, Honglak and Shin, Jinwoo},
  booktitle={Advances in Neural Information Processing Systems},
  pages={7167--7177},
  year={2018}
}

@inproceedings{liu2020energy,
  title={Energy-based out-of-distribution detection},
  author={Liu, Weitang and Wang, Xiaoyun and Owens, John and Li, Yixuan},
  booktitle={Advances in Neural Information Processing Systems},
  volume={33},
  pages={21464--21475},
  year={2020}
}

@article{dong2024multiood,
  title={MultiOOD: Scaling Out-of-Distribution Detection for Multiple Modalities},
  author={Dong, Hao and Zhao, Yue and Chatzi, Eleni and Fink, Olga},
  journal={Advances in Neural Information Processing Systems},
  volume={37},
  year={2024}
}

@article{zhao2019dcso,
  title={DCSO: dynamic combination of detector scores for outlier ensembles},
  author={Zhao, Yue and Hryniewicki, Maciej K},
  journal={arXiv preprint arXiv:1911.10418},
  year={2019}
}

@misc{drori2019automlusingmetadatalanguage,
      title={AutoML using Metadata Language Embeddings}, 
      author={Iddo Drori and Lu Liu and Yi Nian and Sharath C. Koorathota and Jie S. Li and Antonio Khalil Moretti and Juliana Freire and Madeleine Udell},
      year={2019},
      eprint={1910.03698},
      archivePrefix={arXiv},
      primaryClass={cs.LG},
      url={https://arxiv.org/abs/1910.03698}, 
}

@misc{openai2024gpt4technicalreport,
      title={GPT-4 Technical Report}, 
      author={OpenAI and Josh Achiam and Steven Adler and Sandhini Agarwal and Lama Ahmad and others},
      year={2024},
      eprint={2303.08774},
      archivePrefix={arXiv},
      primaryClass={cs.CL},
      url={https://arxiv.org/abs/2303.08774}, 
}

@article{jomaa2021dataset2vec,
  title={Dataset2vec: Learning dataset meta-features},
  author={Jomaa, Hadi S and Schmidt-Thieme, Lars and Grabocka, Josif},
  journal={Data Mining and Knowledge Discovery},
  volume={35},
  number={3},
  pages={964--985},
  year={2021},
  publisher={Springer}
}

@inproceedings{chen2016xgboost,
  title={Xgboost: A scalable tree boosting system},
  author={Chen, Tianqi and Guestrin, Carlos},
  booktitle={Proceedings of the 22nd acm sigkdd international conference on knowledge discovery and data mining},
  pages={785--794},
  year={2016}
}

@misc{fang2024largelanguagemodelsllmstabular,
      title={Large Language Models(LLMs) on Tabular Data: Prediction, Generation, and Understanding -- A Survey}, 
      author={Xi Fang and Weijie Xu and Fiona Anting Tan and Jiani Zhang and Ziqing Hu and Yanjun Qi and Scott Nickleach and Diego Socolinsky and Srinivasan Sengamedu and Christos Faloutsos},
      year={2024},
      eprint={2402.17944},
      archivePrefix={arXiv},
      primaryClass={cs.CL},
      url={https://arxiv.org/abs/2402.17944}, 
}

@article{Li_Ji_Wu_Li_Qin_Wei_Zimmermann_2024, 
title={Panoptic Scene Graph Generation with Semantics-Prototype Learning}, 
volume={38}, url={https://ojs.aaai.org/index.php/AAAI/article/view/28098}, 
DOI={10.1609/aaai.v38i4.28098}, 
number={4}, 
journal={Proceedings of the AAAI Conference on Artificial Intelligence}, 
author={Li, Li and Ji, Wei and Wu, Yiming and Li, Mengze and Qin, You and Wei, Lina and Zimmermann, Roger}, 
year={2024}, 
month={Mar.}, 
pages={3145-3153} }

@inproceedings{li2024dpudynamicprototypeupdating,
  title={Dpu: Dynamic prototype updating for multimodal out-of-distribution detection},
  author={Li, Shawn and Gong, Huixian and Dong, Hao and Yang, Tiankai and Tu, Zhengzhong and Zhao, Yue},
  booktitle={Proceedings of the Computer Vision and Pattern Recognition Conference},
  pages={10193--10202},
  year={2025}
}

@misc{qin2025metaoodautomaticselectionood,
      title={MetaOOD: Automatic Selection of OOD Detection Models}, 
      author={Yuehan Qin and Yichi Zhang and Yi Nian and Xueying Ding and Yue Zhao},
      year={2025},
      eprint={2410.03074},
      archivePrefix={arXiv},
      primaryClass={cs.LG},
      url={https://arxiv.org/abs/2410.03074}, 
}

@misc{ming2022delvingoutofdistributiondetectionvisionlanguage,
      title={Delving into Out-of-Distribution Detection with Vision-Language Representations}, 
      author={Yifei Ming and Ziyang Cai and Jiuxiang Gu and Yiyou Sun and Wei Li and Yixuan Li},
      year={2022},
      eprint={2211.13445},
      archivePrefix={arXiv},
      primaryClass={cs.CV},
      url={https://arxiv.org/abs/2211.13445}, 
}

@misc{wang2023clipnzeroshotooddetection,
      title={CLIPN for Zero-Shot OOD Detection: Teaching CLIP to Say No}, 
      author={Hualiang Wang and Yi Li and Huifeng Yao and Xiaomeng Li},
      year={2023},
      eprint={2308.12213},
      archivePrefix={arXiv},
      primaryClass={cs.CV},
      url={https://arxiv.org/abs/2308.12213}, 
}

@misc{dong2024multioodscalingoutofdistributiondetection,
      title={MultiOOD: Scaling Out-of-Distribution Detection for Multiple Modalities}, 
      author={Hao Dong and Yue Zhao and Eleni Chatzi and Olga Fink},
      year={2024},
      eprint={2405.17419},
      archivePrefix={arXiv},
      primaryClass={cs.CV},
      url={https://arxiv.org/abs/2405.17419}, 
}

@misc{feichtenhofer2019slowfastnetworksvideorecognition,
      title={SlowFast Networks for Video Recognition}, 
      author={Christoph Feichtenhofer and Haoqi Fan and Jitendra Malik and Kaiming He},
      year={2019},
      eprint={1812.03982},
      archivePrefix={arXiv},
      primaryClass={cs.CV},
      url={https://arxiv.org/abs/1812.03982}, 
}

@misc{kay2017kineticshumanactionvideo,
      title={The Kinetics Human Action Video Dataset}, 
      author={Will Kay and Joao Carreira and Karen Simonyan and Brian Zhang and Chloe Hillier and Sudheendra Vijayanarasimhan and Fabio Viola and Tim Green and Trevor Back and Paul Natsev and Mustafa Suleyman and Andrew Zisserman},
      year={2017},
      eprint={1705.06950},
      archivePrefix={arXiv},
      primaryClass={cs.CV},
      url={https://arxiv.org/abs/1705.06950}, 
}

@misc{djurisic2023extremelysimpleactivationshaping,
      title={Extremely Simple Activation Shaping for Out-of-Distribution Detection}, 
      author={Andrija Djurisic and Nebojsa Bozanic and Arjun Ashok and Rosanne Liu},
      year={2023},
      eprint={2209.09858},
      archivePrefix={arXiv},
      primaryClass={cs.LG},
      url={https://arxiv.org/abs/2209.09858}, 
}

@misc{sun2021reactoutofdistributiondetectionrectified,
      title={ReAct: Out-of-distribution Detection With Rectified Activations}, 
      author={Yiyou Sun and Chuan Guo and Yixuan Li},
      year={2021},
      eprint={2111.12797},
      archivePrefix={arXiv},
      primaryClass={cs.LG},
      url={https://arxiv.org/abs/2111.12797}, 
}

@misc{munro2020multimodaldomainadaptationfinegrained,
      title={Multi-Modal Domain Adaptation for Fine-Grained Action Recognition}, 
      author={Jonathan Munro and Dima Damen},
      year={2020},
      eprint={2001.09691},
      archivePrefix={arXiv},
      primaryClass={cs.CV},
      url={https://arxiv.org/abs/2001.09691}, 
}

@INPROCEEDINGS{6126543,
  author={Kuehne, H. and Jhuang, H. and Garrote, E. and Poggio, T. and Serre, T.},
  booktitle={2011 International Conference on Computer Vision}, 
  title={HMDB: A large video database for human motion recognition}, 
  year={2011},
  volume={},
  number={},
  pages={2556-2563},
  doi={10.1109/ICCV.2011.6126543}}

@misc{soomro2012ucf101dataset101human,
      title={UCF101: A Dataset of 101 Human Actions Classes From Videos in The Wild}, 
      author={Khurram Soomro and Amir Roshan Zamir and Mubarak Shah},
      year={2012},
      eprint={1212.0402},
      archivePrefix={arXiv},
      primaryClass={cs.CV},
      url={https://arxiv.org/abs/1212.0402}, 
}

@misc{carreira2018shortnotekinetics600,
      title={A Short Note about Kinetics-600}, 
      author={Joao Carreira and Eric Noland and Andras Banki-Horvath and Chloe Hillier and Andrew Zisserman},
      year={2018},
      eprint={1808.01340},
      archivePrefix={arXiv},
      primaryClass={cs.CV},
      url={https://arxiv.org/abs/1808.01340}, 
}

@misc{dong2023simmmdgsimpleeffectiveframework,
      title={SimMMDG: A Simple and Effective Framework for Multi-modal Domain Generalization}, 
      author={Hao Dong and Ismail Nejjar and Han Sun and Eleni Chatzi and Olga Fink},
      year={2023},
      eprint={2310.19795},
      archivePrefix={arXiv},
      primaryClass={cs.CV},
      url={https://arxiv.org/abs/2310.19795}, 
}

@INPROCEEDINGS{10203747,
  author={Liu, Xixi and Lochman, Yaroslava and Zach, Christopher},
  booktitle={2023 IEEE/CVF Conference on Computer Vision and Pattern Recognition (CVPR)}, 
  title={GEN: Pushing the Limits of Softmax-Based Out-of-Distribution Detection}, 
  year={2023},
  volume={},
  number={},
  pages={23946-23955},
  doi={10.1109/CVPR52729.2023.02293}}

@misc{radford2021learningtransferablevisualmodels,
      title={Learning Transferable Visual Models From Natural Language Supervision}, 
      author={Alec Radford and Jong Wook Kim and Chris Hallacy and Aditya Ramesh and Gabriel Goh and Sandhini Agarwal and Girish Sastry and Amanda Askell and Pamela Mishkin and Jack Clark and Gretchen Krueger and Ilya Sutskever},
      year={2021},
      eprint={2103.00020},
      archivePrefix={arXiv},
      primaryClass={cs.CV},
      url={https://arxiv.org/abs/2103.00020}, 
}

@misc{zhang2023vpgtranstransfervisualprompt,
      title={VPGTrans: Transfer Visual Prompt Generator across LLMs}, 
      author={Ao Zhang and Hao Fei and Yuan Yao and Wei Ji and Li Li and Zhiyuan Liu and Tat-Seng Chua},
      year={2023},
      eprint={2305.01278},
      archivePrefix={arXiv},
      primaryClass={cs.CV},
      url={https://arxiv.org/abs/2305.01278}, 
}

@INPROCEEDINGS{6751553,
  author={Wang, Heng and Schmid, Cordelia},
  booktitle={2013 IEEE International Conference on Computer Vision}, 
  title={Action Recognition with Improved Trajectories}, 
  year={2013},
  volume={},
  number={},
  pages={3551-3558},
  doi={10.1109/ICCV.2013.441}}

@misc{yi2023temporalcoherenttesttimeoptimization,
      title={Temporal Coherent Test-Time Optimization for Robust Video Classification}, 
      author={Chenyu Yi and Siyuan Yang and Yufei Wang and Haoliang Li and Yap-Peng Tan and Alex C. Kot},
      year={2023},
      eprint={2302.14309},
      archivePrefix={arXiv},
      primaryClass={cs.CV},
      url={https://arxiv.org/abs/2302.14309}, 
}

@misc{xu2025legolearnlabelefficientgraphopenset,
      title={LEGO-Learn: Label-Efficient Graph Open-Set Learning}, 
      author={Haoyan Xu and Kay Liu and Zhengtao Yao and Philip S. Yu and Mengyuan Li and Kaize Ding and Yue Zhao},
      year={2025},
      eprint={2410.16386},
      archivePrefix={arXiv},
      primaryClass={cs.LG},
      url={https://arxiv.org/abs/2410.16386}, 
}

@article{li2025secure,
  title={Secure on-device video ood detection without backpropagation},
  author={Li, Shawn and Cai, Peilin and Zhou, Yuxiao and Ni, Zhiyu and Liang, Renjie and Qin, You and Nian, Yi and Tu, Zhengzhong and Hu, Xiyang and Zhao, Yue},
  journal={arXiv preprint arXiv:2503.06166},
  year={2025}
}

@inproceedings{li2023biased,
  title={Biased-predicate annotation identification via unbiased visual predicate representation},
  author={Li, Li and Wang, Chenwei and Qin, You and Ji, Wei and Liang, Renjie},
  booktitle={Proceedings of the 31st ACM International Conference on Multimedia},
  pages={4410--4420},
  year={2023}
}

@inproceedings{li2024domain,
  title={Domain-wise invariant learning for panoptic scene graph generation},
  author={Li, Li and Qin, You and Ji, Wei and Zhou, Yuxiao and Zimmermann, Roger},
  booktitle={ICASSP 2024-2024 IEEE International Conference on Acoustics, Speech and Signal Processing (ICASSP)},
  pages={3165--3169},
  year={2024},
  organization={IEEE}
}

@misc{xu2025glipoodzeroshotgraphood,
      title={GLIP-OOD: Zero-Shot Graph OOD Detection with Graph Foundation Model}, 
      author={Haoyan Xu and Zhengtao Yao and Xuzhi Zhang and Ziyi Wang and Langzhou He and Yushun Dong and Philip S. Yu and Mengyuan Li and Yue Zhao},
      year={2025},
      eprint={2504.21186},
      archivePrefix={arXiv},
      primaryClass={cs.LG},
      url={https://arxiv.org/abs/2504.21186}, 
}

@misc{xu2025fewshotgraphoutofdistributiondetection,
      title={Few-Shot Graph Out-of-Distribution Detection with LLMs}, 
      author={Haoyan Xu and Zhengtao Yao and Yushun Dong and Ziyi Wang and Ryan A. Rossi and Mengyuan Li and Yue Zhao},
      year={2025},
      eprint={2503.22097},
      archivePrefix={arXiv},
      primaryClass={cs.LG},
      url={https://arxiv.org/abs/2503.22097}, 
}
\clearpage
\newpage
\appendix
\section{Pseudo-code for Meta-train and Online Model Selection}

We discussed meta-training and online model selection in Section~\S 3.3, 3.4 and \S~3.5, respectively.
Below are the pseudo-code for the two phases.

\begin{algorithm}
\small
\caption{Offline OOD detection meta-learner training}
\label{alg:offline}
\renewcommand{\algorithmicrequire}
{\textbf{Input:} \parbox[t]{\dimexpr\linewidth-3em}{
    Meta-train database \( \mathcal{D}_{\text{train}} \) composed of \( K \)-modality data,
    model set \( \mathcal{M} \)
}}
\renewcommand{\algorithmicensure}{\textbf{Output:} Meta-learner $f$ for OOD detection model selection}
\begin{algorithmic}[0]
\Require $ $
\Ensure $ $
\State Train and evaluate $\mathcal{M}$ on $\mathcal{D}_{\text{train}}$ to get performance matrix $\mathbf{P}$
\For {$i \in \{1, \dots, n\}$}
    \State Extract data embedding $E^\text{meta}_{i} = \psi(D_{i})= [\psi(x_1), \ldots,  \psi(x_K)]$
    \For {$j \in \{1, \dots, m\}$}
    \State Encode methods set as $E^\text{model}_{j} = \phi(\mathcal{M}, M_j)$
    \State Train $f$ by Eq. (1)
    with the $j$-th model on the $i$-th dataset
    \EndFor
\EndFor
\State \Return the meta-learner $f$
\end{algorithmic}
\end{algorithm}

\begin{algorithm}[H]
\small
\caption{Online OOD detection model selection}
\label{alg:online}
\renewcommand{\algorithmicrequire}{\textbf{Input:} 
 the meta-learner $f$, New ID-OOD dataset pair $D_\text{new}$
}
\renewcommand{\algorithmicensure}{\textbf{Output:} 
 Selected model for $D_\text{new}$
}
\begin{algorithmic}[0]
\Require $ $
\Ensure $ $

\State Extract data embedding, $E^\text{data}_{\text{test}} := \psi(D_\text{new})$
\For {$j \in \{1, \dots, m\}$ (for clarity, written as a for loop)}
\State Encode methods set as $E^\text{model}_{j} = \phi(\mathcal{M}, M_j)$
\State Predict the $j$-th model performance by the meta-learner $f$, i.e., $\widehat{\mathbf{P}}_{\text{new}, j} := f(E^{\text{data}}_{\text{new}}, E^{\text{model}}_j)$
\EndFor

\State \Return the model with the highest predicted perf. by Eq. (2)

\end{algorithmic}
\end{algorithm}

\section{Details of Meta-features}
We introduced the multimodal model embeddings and meta-features for capturing multimodal OOD data characteristics in Section \S 3.4. Table~\ref{tab:meta_all} lists the complete set of meta-features we construct. Part of the meta-features are based on \citep{yi2023temporalcoherenttesttimeoptimization, vanschoren2018meta, 6751553}

\begin{table}[h!]
  \centering
  \renewcommand{\arraystretch}{1.2}
  \scalebox{0.9}{
  \begin{tabular}{@{}lp{0.62\linewidth}@{}}
    \hline
    \textbf{Meta-feature} & \textbf{Definition} \\
    \hline
    \multicolumn{2}{l}{\textit{Video and Optical Flow Related Meta-features}} \\[2pt]
    Clip length $T$                  & Number of RGB frames \\[2pt]
    RGB height $H$                   & Height of each RGB frame in pixels \\[2pt]
    RGB width $W$                    & Width of each RGB frame in pixels \\[2pt]
    RGB aspect ratio                 & $H/W$ \\[2pt]
    Flow height $H'$                 & Height of each optical-flow frame \\[2pt]
    Flow width $W'$                  & Width of each optical-flow frame \\[2pt]
    Flow aspect ratio                & $H'/W'$ \\[2pt]
    Colourfulness index              & Hasler--Süsstrunk measure computed from all pixels \\[2pt]
    Edge density                     & Fraction of Canny edge pixels, averaged over time \\[2pt]
    GLCM entropy                     & Average grey-level co-occurrence entropy over frames \\[2pt]
    HoF histogram                    & Eight-bin, magnitude-weighted histogram of flow orientations covering $(-\pi,\pi]$ \\[2pt]
    \hline
    \multicolumn{2}{l}{\textit{Basic Statistics}} \\[2pt]
    $\mu_I$ (Clip mean)            & Mean value \\[2pt]
    $\sigma_I$ (Clip std)          & Standard deviation value\\[2pt]
    $\mathrm{skew}_I$              & Skewness of distribution \\[2pt]
    $\mathrm{kurt}_I$              & Kurtosis of distribution \\[2pt]
    $\min_I$                       & Minimum value\\[2pt]
    $\max_I$                       & Maximum value\\[2pt]
    $\mathrm{med}_I$               & Median value\\[2pt]
    $\mathrm{IQR}_I$               & Interquartile range of intensities \\[2pt]
    $\mathrm{Gini}_I$              & Gini coefficient of values \\[2pt]
    $\mathrm{MAD}_I$               & Median absolute deviation value \\[2pt]
    $\mathrm{AAD}_I$               & Mean absolute deviation value \\[2pt]
    $\mathrm{CV}_I$                & Coefficient of variation (std/mean) \\[2pt]
    $p_{\mathrm{out},I}^{1\%}$     & Proportion outside 1st–99th percentile\\[2pt]
    $p_{\mathrm{out},I}^{3\sigma}$ & Proportion outside $\mu_I\pm3\sigma_I$\\[2pt]
    $\mu_M$ (Flow mean)            & Mean of optical-flow magnitudes \\[2pt]
    $\sigma_M$ (Flow std)          & Standard deviation of optical-flow magnitudes \\[2pt]
    $\mathrm{IQR}_M$               & Interquartile range of flow magnitudes \\[2pt]
    $p_{\mathrm{out},M}^{1\%}$     & Proportion outside 1st–99th percentile of flow magnitudes \\[2pt]
    $p_{\mathrm{out},M}^{3\sigma}$ & Proportion outside $\mu_M\pm3\sigma_M$ of flow magnitudes \\
    \hline
  \end{tabular}}
  \caption{Details of the meta-features. Meta-features include CLIP-based per-frame meta-features and the grayscale and motion meta-features.}
  \label{tab:meta_all}
\end{table}

\section{ Details of Baselines}
Section~\S4 introduces the model selection baselines, which we choose based on prior work in meta-learning for unsupervised model selection \citep{NEURIPS2021_23c89427,10027699,jiang2024adgym,park2023metagl}. These baselines are grouped into four categories, as shown in main Tab.~4. Further details are provided  below:

\textbf{(a) No model selection or random selection}: always employs either the ensemble of all the models or the same single model, or randomly selects a model:
  \textbf{\textit{(1) Maximum Softmax Probability (MSP)} }\citep{hendrycks17baseline}, a popular OOD detector that uses the maximum softmax score of a neural network's logits as threshold to identify whether an input belongs to the distribution the network was trained on.
  \textbf{\textit{(2) Mahalanobis (MD)}} \citep{lee2018simple} computes the distance between a sample’s features and class means using the Mahalanobis metric, treating lower distances as more likely to be in-distribution.
   \textbf{\textit{(3) Mega Ensemble (ME)}} averages OOD scores from all the models for a given dataset without performing model selection but rather using \textit{all} the models.
   \textbf{\textit{(4) Random Selection (Random)}}
   selects a model at random from the set of available candidate detectors.

\textbf{(b) Simple meta-learners} that do not involve optimization:
  \textbf{\textit{(5) Global Best (GB)}} is the \textit{simplest meta-learner} that selects the model with the largest average performance across all meta-train datasets. GB does {\em not} use any meta-features.
  \textbf{\textit{(6) ISAC}}~\citep{conf/ecai/KadiogluMST10} clusters the meta-train datasets based on meta-features. Given a new dataset pair, it identifies its closest cluster and selects the best-performing model in that cluster.
  \textbf{\textit{(7) ARGOSMART (AS)}}~\citep{nikolic2013simple} finds the closest meta-train dataset (1 nearest neighbor) to a given test datasetin terms of meta-feature distance, and selects the model with the best performance on the 1NN dataset.

\textbf{(c) Optimization-based meta-learners} which involves a learning process:
  \textbf{\textit{(8) ALORS}}\citep{journals/ai/MisirS17} factorizes the performance matrix to extract latent factors and estimate the performance matrix as the dot product of the latent factors. A regressor maps meta-features to these latent factors.
  \textbf{\textit{(9) NCF}}~\citep{10.1145/3038912.3052569} replaces the dot product used in ALORS with a more general neural architecture that predicts performance by combining the linearity of matrix factorization and non-linearity of deep neural networks.

\textbf{(d) Large language models (LLMs) as a model selector}:
\textbf{\textit{(10) GPT-4o mini}} \citep{openai2024gpt4technicalreport} used as zero-shot meta-selector. The dataset and method descriptions are directly provided to the LLM, allowing it to select the methods based on these descriptions. Note there is no meta-learning here. Details are in Supplementary Material \S 3.

\subsection{Random Selection}
The Random Selection baseline uses random seed 42.

\subsection{GPT-4o-mini}
GPT-4o-mini \citep{openai2024gpt4technicalreport} is used as one of the baselines, serving as a zero-shot meta-selector. The text inputs are as follows:

\subsubsection{Datasets Descriptions}\label{sec:datsaet}
\textbf{EPIC-Kitchens}: A large-scale egocentric dataset collected by 32 participants recording daily kitchen activities in their homes. We use the subset from the Multimodal Domain Adaptation benchmark, comprising 4,871 clips from the 8 largest action classes in sequence P22: put, take, open, close, wash, cut, mix, and pour. Modalities include video, optical flow, and audio.

\textbf{HAC}: Contains 3,381 video clips of 7 actions: sleeping, watching TV, eating, drinking, swimming, running, and opening door, performed by humans, animals, and cartoon figures. Includes video, optical flow, and audio modalities.

\textbf{UCF101}: Comprises 13,320 YouTube video clips covering 101 actions with significant diversity in motion, appearance, and background. Modalities include video and optical flow.

\textbf{Kinetics-600}: A large-scale dataset of \~480k 10-second clips spanning 600 actions. We use a subset of 229 classes (57,205 clips) to reduce class overlap with other datasets. Optical flow is extracted at 24 FPS using the TV-L1 algorithm, totaling 114,410 samples. Final modalities include video, audio, and optical flow.

\subsubsection{OOD Detector Descriptions}
\textbf{MSP}: Implements the Maximum Softmax Probability (MSP) Thresholding baseline for OOD detection.

\textbf{EnergyBased}: Calculates the negative energy for a vector of logits. This value is used as the outlier score.

\textbf{MaxLogit}: Implements the Max Logit Method for OOD Detection, as proposed in Scaling Out-of-Distribution Detection for Real-World Settings.

\textbf{Mahalanobis}: Calculates a class center for each class and a shared covariance matrix from the data. It also uses ODIN preprocessing.

\textbf{ReAct}: Clips the activations in some layer of the network (backbone) and forward propagates the result through the remainder of the model (head). In the paper, ReAct is applied to the penultimate layer of the network. The output of the network is then passed to an outlier detector that maps the output of the model to outlier scores.

\textbf{ASH}: Prunes the activations in some layer of the network (backbone) by removing a certain percentile of the highest activations. The remaining activations are modified, depending on the particular variant selected, and propagated through the remainder (head) of the network. Then uses the energy-based outlier score. This approach has been shown to increase OOD detection rates while maintaining ID accuracy.

\textbf{GEN}: Utilizes the entropy of the softmax output as a measure of confidence. In-distribution samples are expected to have higher confidence (lower entropy), while OOD samples will exhibit lower confidence (higher entropy).

\textbf{ViM}: Implements Virtual Logit Matching (ViM) from the paper ViM: Out-Of-Distribution Detection with Virtual-logit Matching.

\textbf{KNN}: Implements the detector from the paper Out-of-Distribution Detection with Deep Nearest Neighbors. Fits a nearest neighbor model to the IN samples and uses the distance from the nearest neighbor as the outlier score.

\subsubsection{Prompt}
The prompt provided to the LLM is structured as follows, with text descriptions of both the datasets and models provided.
To ensure consistency, we set \textit{temperature} parameter to 0, and \textit{top\_p} parameter to 0.999. 

\texttt{
[Dataset descriptions provided]}

\texttt{
Your task is to select the best OOD detection method for a set of 24 test ID-OOD dataset pairs. 
You will be provided with descriptions of both the ID-OOD dataset pairs and the available OOD detection methods. 
You should pick the best model that has the highest AUROC metric. 
For each dataset pair, output the recommended OOD detection method in the format: 'Recommended Method: [Recommended Method]'.
}

\texttt{
[Model descriptions provided]
}

\section{Near-OOD and Far-OOD Setup}
M3OOD utilizes five video datasets comprising more than 85,000 video clips in total. These datasets differ in the number of classes, which range from 7 to 229, and in size, ranging from 3,000 to 57,000 clips. Video and optical flow are used as different types of modalities. Details of the five datasets are in \S~3.1.1.

In the \textbf{Near-OOD} setup, four datasets are used. EPIC-Kitchens 4/4 is derived from the EPIC-Kitchens Domain Adaptation dataset \citep{munro2020multimodaldomainadaptationfinegrained}, with four classes used for training as ID and four for testing as OOD, totaling 4,871 video clips. HMDB51 25/26 and UCF101 50/51 are constructed from HMDB51 \citep{6126543} and UCF101 \citep{soomro2012ucf101dataset101human}, with 6,766 and 13,320 clips, respectively. Kinetics-600 129/100 uses 229 classes selected from Kinetics-600 \citep{carreira2018shortnotekinetics600}, each with about 250 clips (57,205 total); 129 classes are used as ID and 100 as OOD.

In the \textbf{Far-OOD} setup, HMDB51 and Kinetics-600 are used as ID datasets. For HMDB51 as ID, OOD datasets include UCF101, EPIC-Kitchens, HAC, and Kinetics-600. To avoid class overlap, we exclude 8 overlapping classes from HMDB51 (leaving 43 ID classes) and remove 31 overlapping classes from UCF101 (resulting in 70 OOD classes). For the remaining datasets, no overlap exists and their original classes are used.
For Kinetics-600 as ID, the same OOD datasets are adopted. We use the same 229-class subset from the Near-OOD setup to reduce overlap. For UCF101, 11 overlapping classes are removed, leaving 90 classes as OOD. Other datasets are used as-is due to no class overlap.

\section{Additional Experiment Setting}
We select the parameters for M3OOD and M3OOD\_NN (used in the ablation study) through grid search. The final parameter configurations are provided in the code repo.

\section{Additional Results}
\begin{figure}[h]
    \centering
\includegraphics[width=\linewidth]{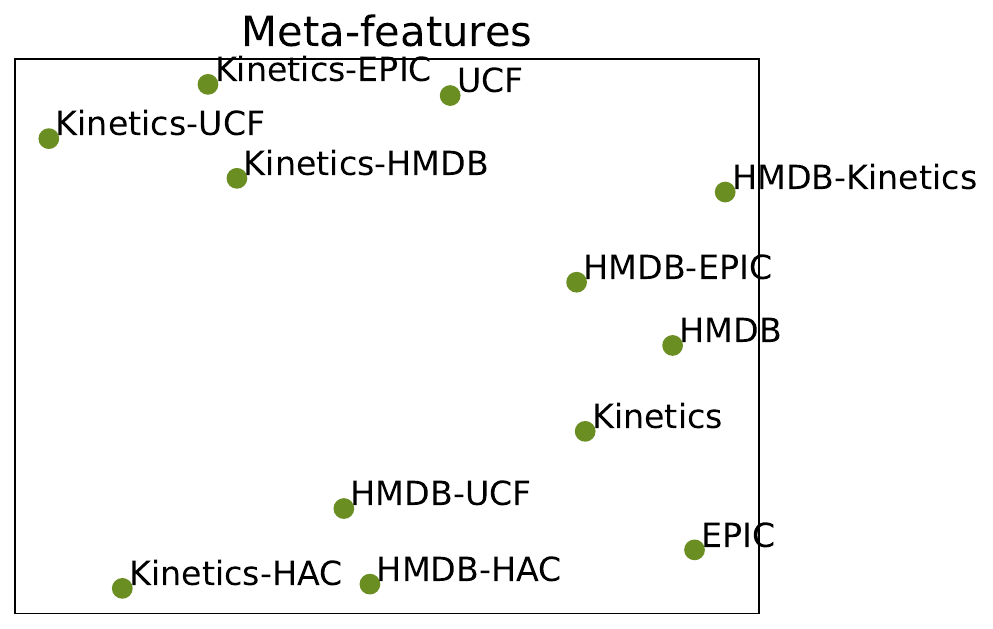}
    \caption{Visualization of dataset embeddings.}
    \label{fig:tsne}
\end{figure}

Figure~\ref{fig:tsne} shows the dataset embeddings visualization, with the embeddings reduced to 2D using t-SNE.
We observe clear clustering patterns that reflect underlying similarities across datasets. For instance, datasets originating from the same source or sharing overlapping label spaces, such as Kinetics-HMDB and Kinetics-UCF, are located closely, indicating that the meta-features capture alignment in distribution or content. Similarly, HMDB-EPIC and HMDB-Kinetics are proximal to HMDB, suggesting consistency in the extracted features when paired with other datasets
Moreover, datasets involving HAC (e.g., Kinetics-HAC, HMDB-HAC) appear in a distinct region, separated from others. This spatial distinction implies that the HAC dataset exhibits different properties—such as lower visual diversity, temporal resolution, or action granularity—compared to datasets like Kinetics and EPIC. This separation also highlights the ability of the meta-features to reflect meaningful dataset differences relevant for model selection and generalization.

\section{Details on Notations}
The following notations are used in Main Fig.~1 and Main Alg.~2 for comprehensive M3OOD overview.
\begin{table}[h]
\small
    \begin{tabular}{ll}
    \hline
         \textbf{Notations} & \textbf{Description} \\
         \hline
         $\mathcal{L}$ & Training Loss \\
         $\mathcal{M}$ & \# OOD Detection Methods \\
         $\mathcal{N}$ & \#  Dataset Pairs \\
         $D$ & Datast Pair \\
         $\phi$ & Embedding Notation for OOD Detectors \\
         $\psi$ & Embedding Notation for Dataset Pairs \\
         $E$ & Embeddings for datasets and models\\
         $P_{i,j}$ & Performance of OOD Detector $j$ on \\
          &  Dataset Pair $i$ \\
         $\hat{P}_{i,j}$ & Predicted Performance of OOD Detector $j$\\
          &  on Dataset Pair $i$ \\
         $f$ & Meta-predictor \\
         \hline
    \end{tabular}
    \caption{Notations with details used in Main Fig.~1 and Alg.~2.} 
    \label{table:notations}
\end{table}
\end{document}